\documentclass[conference]{IEEEtran}
\IEEEoverridecommandlockouts
\usepackage{cite}
\usepackage{amsmath,amssymb,amsfonts}
\usepackage{algorithmic}
\usepackage{graphicx}
\usepackage{textcomp}
\usepackage{xcolor}
\usepackage{multirow}
\usepackage{array}
\usepackage{url}
\usepackage[bottom]{footmisc} % 固定脚注在页面底部
\usepackage{makecell}
\usepackage{booktabs}

\def\BibTeX{{\rm B\kern-.05em{\sc i\kern-.025em b}\kern-.08em
    T\kern-.1667em\lower.7ex\hbox{E}\kern-.125emX}}
\makeatletter
\def\@IEEENORMtitlevspace{1.0\baselineskip} % 原本可能是 1.5 或 2.0
\def\@IEEEMINtitlevspace{0.5\baselineskip}
\makeatother

\begin{document}

\title{BiLCNet : BiLSTM-Conformer Network\\ for Encrypted Traffic Classification \\ with 5G SA Physical Channel Records}

\author{\IEEEauthorblockN{Ke Ma}
\IEEEauthorblockA{\textit{SJTU Paris Elite Institute of Technology} \\
\textit{Shanghai Jiao Tong University}\\
Shanghai, China \\
vanessa0717@sjtu.edu.cn}
\and
\IEEEauthorblockN{Jialiang Lu}
\IEEEauthorblockA{\textit{SJTU Paris Elite Institute of Technology} \\
\textit{Shanghai Jiao Tong University}\\
Shanghai, China \\
jialiang.lu@sjtu.edu.cn}
\and
\IEEEauthorblockN{Philippe Martins}
\IEEEauthorblockA{\textit{Telecom Paris} \\
\textit{Institut Polytechnique de Paris}\\
Palaiseau, France \\
martins@telecom-paris.fr}
% \and
% \and
% \and
}
\maketitle
\begin{abstract}
Accurate and efficient traffic classification is vital for wireless network management, especially under encrypted payloads and dynamic application behavior, where traditional methods such as port-based identification and deep packet inspection (DPI) are increasingly inadequate. This work explores the feasibility of using physical channel data collected from the air interface of 5G Standalone (SA) networks for traffic sensing. We develop a preprocessing pipeline to transform raw channel records into structured representations with customized feature engineering to enhance downstream classification performance. To jointly capture temporal dependencies and both local and global structural patterns inherent in physical channel records, we propose a novel hybrid architecture: BiLSTM-Conformer Network (BiLCNet), which integrates the sequential modeling capability of Bidirectional Long Short-Term Memory networks (BiLSTM) with the spatial feature extraction strength of Conformer blocks. Evaluated on a noise-limited 5G SA dataset, our model achieves a classification accuracy of 93.9\%, outperforming a series of conventional machine learning and deep learning algorithms. Furthermore, we demonstrate its generalization ability under zero-shot transfer settings, validating its robustness across traffic categories and varying environmental conditions.
\end{abstract}


\section{Introduction}\label{A}
Traffic classification is a fundamental task in modern mobile networks, enabling effective resource allocation, content-aware billing, access control, and anomaly detection. Its performance directly affects quality of service (QoS), network security, and management efficiency\cite{azab2024network}. However, the widespread adoption of encryption in services such as e-commerce, social media, and video streaming has made accurate traffic identification increasingly challenging\cite{wang2019survey}. Traditional methods such as port-based approaches and deep packet inspection (DPI) are no longer sufficient, highlighting the need for advanced learning-based techniques capable of handling encrypted and diverse traffic in real time.

\subsection{Traditional Traffic Classification Techniques}

Port-based identification is the oldest traffic classification method that maps known Transmission Control Protocol/User Datagram Protocol (TCP/UDP) ports to services. While simple and efficient, its reliability has declined due to dynamic ports, tunneling, and Network Address Port Translation (NAPT)\cite{madhukar2006longitudinal}.

Another traditional approach is DPI, which analyzes packet headers and payloads using pattern matching to identify applications. While it achieves high accuracy for unencrypted traffic, its high computational cost and poor performance under encryption significantly limit its practical applicability\cite{finsterbusch2013survey}.

\subsection{Learning-based Traffic Classification Techniques}

Learning-based methods, mainly including machine learning (ML) and deep learning (DL) algorithms, have emerged to address the limitations of traditional approaches under encryption. Representative learning-based methods are reviewed in the following, with key approaches and their challenges summarized in Table \ref{tab:traffic_classification_summary}.

Supervised learning methods such as K-Nearest Neighbors (KNN)\cite{roughan2004class}, Naive Bayes\cite{moore2005internet}, and Support Vector Machines (SVM) have shown promise in traffic classification but face key limitations: data distribution sensitivity, dependence on labeled data, and limited real-time use. In contrast, unsupervised approaches like K-means\cite{bernaille2006traffic} and Expectation-Maximization (EM)\cite{zander2005automated} remove the need for labeled data but often suffer from unstable performance and rigid clustering structures, leading to reduced accuracy for complex traffic patterns. In addition, semi-supervised learning\cite{glennan2016improved} has also been widely explored, though it typically requires more computational resources than both supervised and unsupervised methods.

In recent years, deep learning has become a mainstream approach for encrypted traffic classification, with Convolutional Neural Networks (CNNs)\cite{o2016convolutional}, Graph Neural Networks (GNNs)\cite{pang2021cgnn}, and Recurrent Neural Networks (RNNs)\cite{rajendran2018deep} widely used to extract temporal features. More recently, Transformer-based models have shown strong capabilities in traffic analysis due to their attention mechanisms and ability to capture long-range dependencies. There have also been various attempts to combine different architectures, such as integrating CNN with LSTM\cite{wang2017hast} or fusing CNN and Transformer models\cite{liu2025transeca}. However, they often suffer from high computational costs, and limited generalization remains a key challenge for real-time sensing in 5G networks.

\begin{table*}[ht]
\centering
\caption{Comparison of Learning-based Techniques in Traffic Classification}
\renewcommand{\arraystretch}{1.2}
\begin{tabular}{|c|c|c|>{\centering\arraybackslash}p{6.6cm}|}
\hline
\textbf{Technique} & \textbf{Algorithm} & \textbf{Input Features} & \textbf{Main Challenge} \\
\hline
\multirow{2}{*}{Supervised Learning}
& KNN\cite{roughan2004class} & Network traffic statistical features & Poor distinction between streaming and bulk flows \\
& Naive Bayes\cite{moore2005internet} & TCP/UDP flow statistical features & Strong dependence on dataset distribution \\
\hline
\multirow{2}{*}{Unsupervised Learning}
& K-means\cite{bernaille2006traffic} & TCP flow packet-level features & High sensitivity to initial cluster center selection \\
& EM\cite{zander2005automated} & Network traffic statistical features & Low separability with similar feature distributions \\
\hline
\multirow{5}{*}{Deep Learning}
& CNN\cite{o2016convolutional} & Raw radio time series & Reduced classification accuracy at low SNR \\
& GNN\cite{pang2021cgnn} & Network traffic graph features & Structural limitations with increasing depth \\
& RNN\cite{rajendran2018deep} & Time-domain amplitude and phase & Reduced classification accuracy at low SNR \\
& CNN + LSTM\cite{wang2017hast} & Network traffic statistical features & Poor performance on imbalanced datasets \\
& CNN + Transformer\cite{liu2025transeca} & Network traffic statistical features & Limited generalization to zero-day traffic patterns \\
\hline
\end{tabular}
\label{tab:traffic_classification_summary}
\end{table*}

To address the above limitations, we propose BiLCNet, a hybrid BiLSTM-Conformer model for encrypted traffic classification. The principal contributions of this work include:
\begin{itemize}
\item We design BiLCNet, which integrates Bidirectional LSTM for temporal modeling, CNN for local feature extraction, and Transformer architecture for global attention to improve classification performance.
\item We utilize 5G physical channel data from the wireless air interface - an underutilized yet highly informative data source collected under diverse channel conditions, and design a preprocessing pipeline with feature engineering to extract informative representations.
\item Unlike models designed for real-time classification, our model prioritizes high accuracy and strong generalization in offline scenarios across diverse unseen traffic.
\end{itemize}

\section{Preliminaries}
\subsection{5G Mobile Communication Network Architecture}
\begin{figure}[b]
    \centering
    \includegraphics[width=1\linewidth]{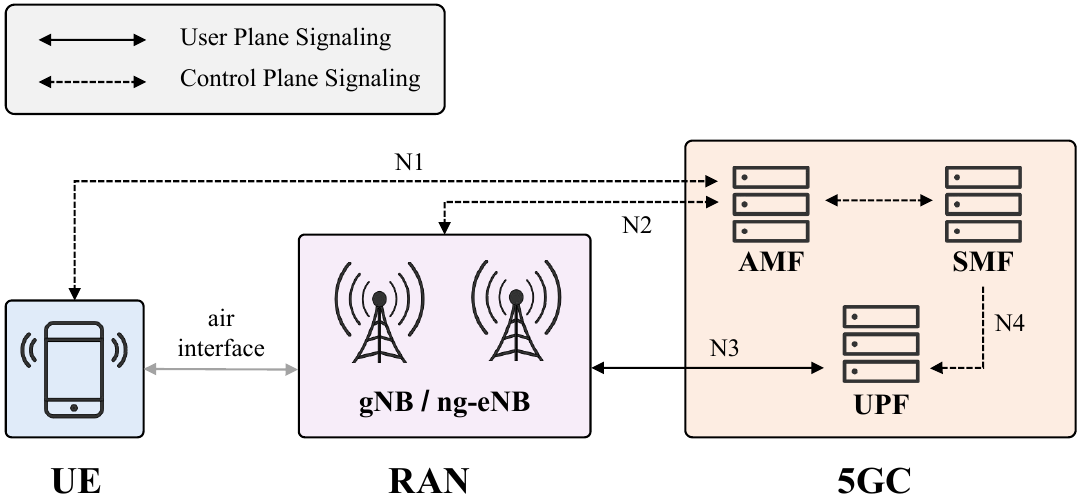}
    \caption{5G Mobile Communication Network Architecture}
    \label{fig:1}
\end{figure}
The 5G mobile communication network represents a significant upgrade from Fourth Generation Long-Term Evolution (4G LTE), offering higher speed, lower latency, and greater capacity. As shown in Fig.~\ref{fig:1}, its architecture consists of three core components: the User Equipment (UE), such as smartphones and IoT devices; the Radio Access Network (RAN), which manages wireless connections and allocates radio resources; and the 5G Core Network (5GC), responsible for overall control and data routing. Wireless links connect the UE and RAN, while the RAN and core network communicate through control and user plane interfaces.

The simulation platform used to generate the dataset in this study is built upon this standard 5G architecture, specifically adopting the Standalone (SA) deployment mode. Unlike the Non-Standalone (NSA) architecture, which integrates 5G radio access with the existing 4G LTE infrastructure, SA operates independently, employing the 5GC and gNodeB (gNB) as its core and access components, respectively. Selecting the SA deployment for our experimental platform results in a simplified network stack and more faithfully represents native 5G application scenarios.

\subsection{Physical Channels of the 5G Wireless Air Interface}\label{AA}
In 5G networks, wireless communication between the UE and gNB is facilitated through three types of channels: logical channels, transport channels, and physical channels. In this study, we primarily focus on physical channel data from the 5G wireless air interface. This data provides valuable insights without requiring packet decryption, thereby enhancing privacy and reducing computational overheads. Moreover, its inherited features from upper layers allow it to promptly reflect service-level variations, which is particularly beneficial for accurate and timely traffic classification. Although this approach has not been widely explored, it holds promise for advancing traffic classification techniques.

Specifically, 5G physical channels operate within the physical (PHY) layer and are responsible for the actual transmission of signals over radio resources. There are six standardized physical channels in 5G: the Physical Uplink/Downlink Control Channel (PUCCH/PDCCH), Physical Uplink/Downlink Shared Channel (PUSCH/PDSCH), Physical Broadcast Channel (PBCH), and Physical Random Access Channel (PRACH). Their transmission content is summarized in Table \ref{tab:phy-channels}.

\begin{table}[b]
\centering
\caption{Summary of 5G Physical Channels}
\renewcommand{\arraystretch}{1.2}
\begin{tabular}{|c|c|>{\centering\arraybackslash}p{4.2cm}|}
\hline
\textbf{Channel} & \textbf{Direction} & \textbf{Transmission Content} \\
\hline
PDCCH & Downlink & Physical Layer Control information \\
PDSCH & Downlink & User data \\
PBCH  & Downlink & System information for access \\
PUCCH & Uplink   & Physical Layer Control information \\
PUSCH & Uplink   & User data \\
PRACH & Uplink   & Signals for random access attempts \\
\hline
\end{tabular}
\label{tab:phy-channels}
\end{table}

\section{Methodology}
\subsection{Data Collection}
Our 5G physical channel dataset was collected in a Faraday cage at the INFRES\footnote{Department of Computer Science and Networks, Télécom Paris} laboratory to eliminate external electromagnetic interference. The experimental platform simulates a realistic 5G communication environment by integrating the functions of both the RAN and the 5GC on a single computer running the Amarisoft\footnote{Amarisoft official website: \url{https://www.amarisoft.com/}} 5G protocol stack. This computer is connected to a software-defined radio (SDR) transceiver, which handles the transmission and reception of wireless signals. On the user side, a commercial 5G smartphone serves as the UE, establishing a wireless air interface link with the SDR. This setup allows us to simulate various user behaviors on the UE while capturing the corresponding physical-layer log data through Amarisoft’s data acquisition software.

Overall, we collected 5G physical channel data across four distinct user behavior patterns and eleven different channel conditions. The behavior patterns considered were (1) voice calls, (2) video meetings, (3) large file uploads, and (4) large file downloads. To simulate varying wireless channel qualities, the transmission gain of the access network was adjusted from 64 dB to 84 dB in increments of 2 dB. Higher transmission gain values correspond to better channel conditions.

\subsection{Data Preprocessing}
\begin{figure}[b]
    \centering
    \includegraphics[width=1\linewidth]{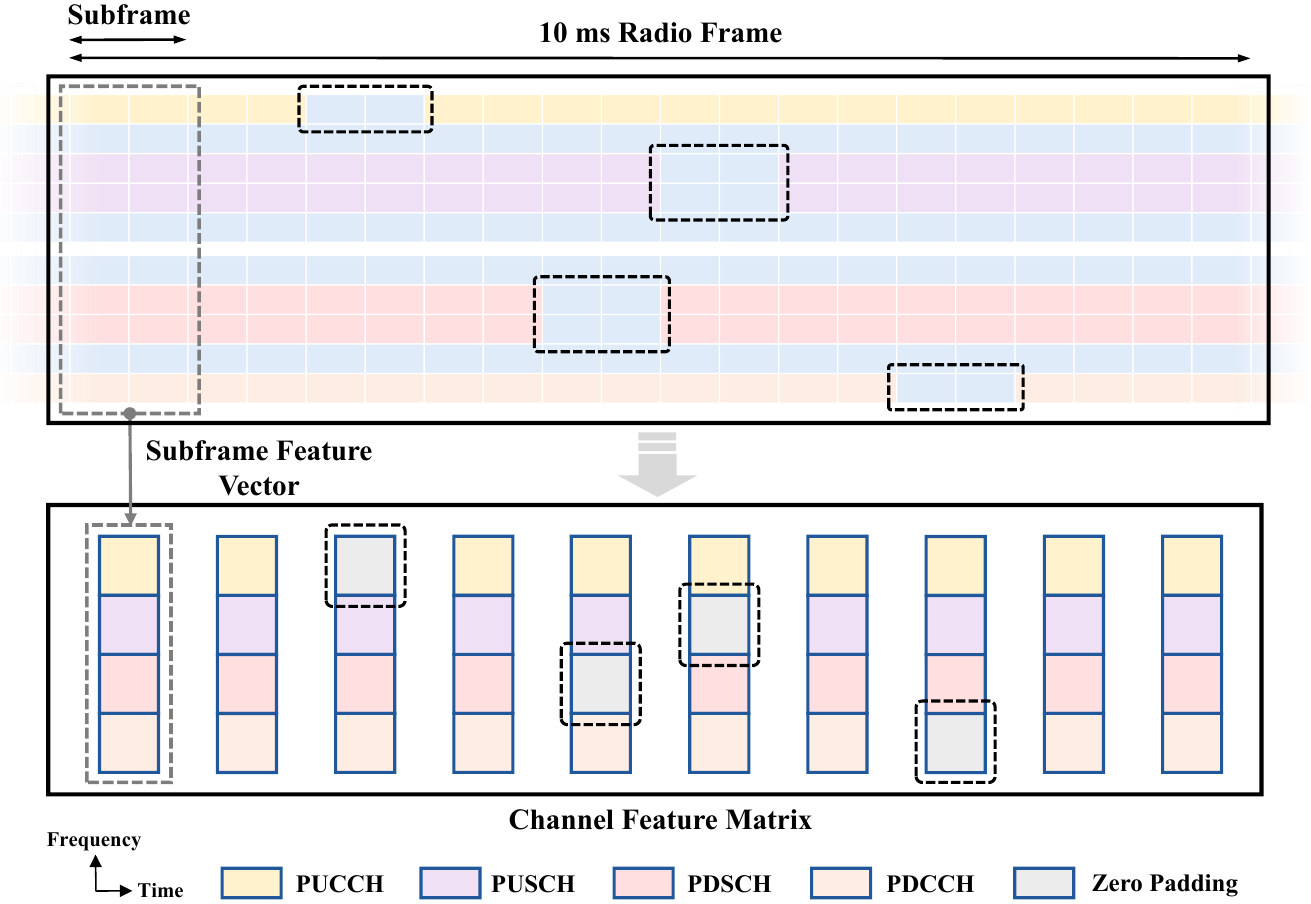}
    \caption{Data Preprocessing of 5G SA Physical Channel Features ($\mu = 1$)}
    \label{fig:preprocessing}
\end{figure}
The raw dataset consists of log files containing physical channel feature data. We design a data preprocessing pipeline based on the 5G SA frame structure to transform these logs into structured, trainable data representations, inspired by \cite{fei2023real}, as illustrated in Fig.~\ref{fig:preprocessing}.

5G employs orthogonal frequency-division multiple access (OFDMA) to divide bandwidth into interference-free subcarriers for fine-grained resource allocation. In 5G SA systems, the physical layer adopts a fixed radio frame length of 10\,ms, with each frame comprising 10 subframes of 1\,ms. For numerology $\mu = 1$, each subframe is further divided into 2 slots, and each time-frequency unit corresponds to a resource block (RB). 

Based on this frame structure, the preprocessing pipeline operates as follows. In a 10\,ms frame-based observation window, for each subframe, we extract features from four key physical channels: PUCCH, PUSCH, PDSCH, and PDCCH, and concatenate them into a fixed-length Subframe Feature Vector. Zero-padding is applied if a channel is missing. The feature vectors of all 10 subframes are then stacked to form a Channel Feature Matrix, providing a compact and consistent input representation for downstream learning tasks.

The dataset includes measurements from four physical channels, but many features appear only occasionally in the records, making them inefficient for deep learning models. To improve computational performance, we select over 60 key features that are most relevant to traffic classification. The key variables are summarized in the table \ref{tab:feature_meanings}.

\begin{table}[h]
\centering
\caption{Physical Meaning of Key Selected Features}
\renewcommand{\arraystretch}{1.2}
\begin{tabular}{|>{\centering\arraybackslash}p{1.5cm}|
                >{\centering\arraybackslash}p{1.2cm}|
                p{4.3cm}|}
\hline
\textbf{Category} & \textbf{Feature} & \multicolumn{1}{c|}{\textbf{Physical Meaning}} \\
\hline
\multirow{5}{*}{\makecell[c]{User Data\\Stream}}
    & \textit{mcs}     & Modulation and Coding Scheme \\
    & \textit{harq}    & Hybrid Automatic Repeat Request \\
    & \textit{tb\_len} & Length of Transport Block \\
    & \textit{prb}     & Physical Resource Block Allocation \\
    & \textit{symb}    & Symbol Allocation Position \\
\hline
\multirow{3}{=}{Air Interface Environment} 
    & \textit{snr}     & Signal-to-Noise Ratio at Receiver \\
    & \textit{epre}    & Energy per Resource Element \\
    & \textit{cce}     & Control Channel Element \\
\hline
\end{tabular}
\label{tab:feature_meanings}
\end{table}

Following the extraction of the aforementioned features, additional \textbf{feature engineering}\cite{aphayavong2024optimizing} techniques are employed to improve the model’s generalization capability. In particular, a range of derived statistical descriptors is computed to reveal implicit relationships within the data.

\textit{1) Cumulative Error Rate}: The cumulative uplink/downlink error rate (\( \mathrm{ERR} \)) is calculated based on Hybrid Automatic Repeat Request (HARQ) feedback as follows:
\begin{equation}
\mathrm{ERR} = 1 - \frac{N_{\mathrm{succ}}}{N_{\mathrm{total}}}
\label{eq:err}
\end{equation}
where \( N_{\mathrm{succ}} \) is the number of successful HARQ transmissions and \( N_{\mathrm{total}} \) the total HARQ attempts. A higher \( \mathrm{ERR} \) means less reliable traffic type needing stronger error correction, while a lower \( \mathrm{ERR} \) indicates more stable transmission.

\textit{2) PDSCH Efficiency}: The efficiency of PDSCH is defined as the ratio between the total transmitted transport block size and the total number of Physical Resource Blocks (PRBs) allocated:

\begin{equation}
\mathrm{EFF}_{\mathrm{PDSCH}} = \frac{\sum TB_{\mathrm{PDSCH}}}{\sum \mathrm{PRB}_{\mathrm{PDSCH}}}
\label{eq:pdsch_efficiency}
\end{equation}

A higher \( \mathrm{EFF}_{\mathrm{PDSCH}} \) indicates more efficient utilization of downlink resources, typically associated with real-time or latency-sensitive services. 

\begin{figure*}[ht]
    \centering
    \includegraphics[width=1\linewidth]{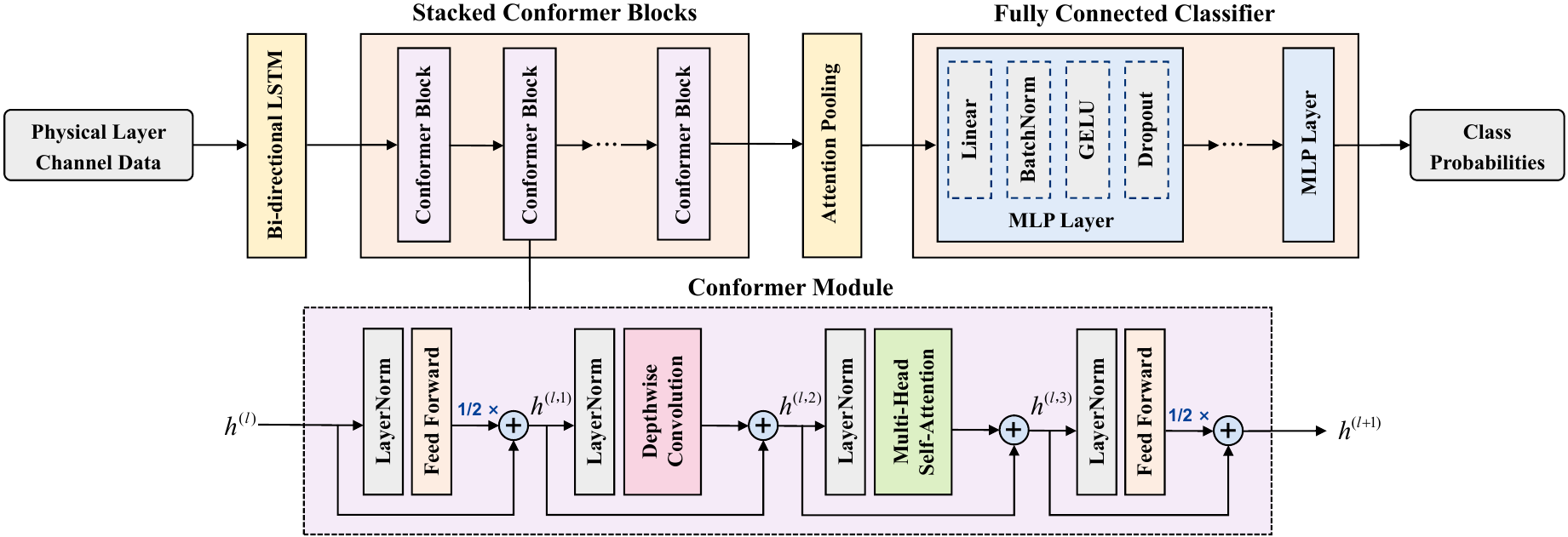}
    \caption{Architecture of BiLCNet}
    \label{fig:model}
\end{figure*}

\textit{3) Modulation Variability Index}: The Modulation Variability Index (MVI) quantifies the fluctuation in modulation schemes within a given observation window. It is defined as the coefficient of variation of the modulation order:

\begin{equation}
\mathrm{MVI} = \frac{\sigma_{\mathrm{mod}}}{\mu_{\mathrm{mod}}}
\end{equation}
where \( \sigma_{\mathrm{mod}} \) and \( \mu_{\mathrm{mod}} \) denote the standard deviation and mean of modulation order values, respectively. A lower MVI indicates stable modulation, while a higher MVI reflects greater variability due to adaptive modulation in dynamic conditions.

\subsection{BiLCNet : BiLSTM-Conformer Network}
Through data preprocessing, we construct frame-structured channel feature matrices from physical-layer logs. Each subframe is encoded as a feature vector, with ten vectors stacked to form a matrix. This representation encodes both temporal dependencies across subframes and spatial correlations within each subframe. Therefore, to effectively exploit these characteristics, we design BiLCNet, a hybrid model illustrated in Fig.~\ref{fig:model}. It combines a bidirectional LSTM for temporal modeling, while the Conformer blocks jointly learn spatial correlations and temporal dependencies through convolution and self-attention.

\textbf{Bidirectional LSTM Layer} To model temporal dependencies along the sequence, the input feature matrix \( \mathbf{X}_t \in \mathbb{R}^{T \times D} \), where \( T \) is the number of subframes and \( D \) is the feature dimension of each subframe, is first fed into a bidirectional LSTM (BiLSTM) with \( L \) layers and hidden dimension \( H \). The BiLSTM processes sequential data in both forward and backward directions to capture contextual information. At each time step $t$, the hidden states $\mathbf{H}_t\in \mathbb{R}^{n \times 2H}$ are computed as:

\begin{equation}
\mathbf{H}_t = \left[\text{LSTM}^{(f)}(\mathbf{X}_t, \overrightarrow{\mathbf{H}}_{t-1}); \text{LSTM}^{(b)}(\mathbf{X}_t, \overleftarrow{\mathbf{H}}_{t+1})\right]
\end{equation}
where $n$ denotes the batch size, $\text{LSTM}^{(f)}$ and $\text{LSTM}^{(b)}$ denote the forward and backward LSTM operations respectively, and $[\cdot;\cdot]$ represents concatenation. The final output $\mathbf{O}_t\in \mathbb{R}^{T \times 2H}$ is obtained through a linear transformation of $\mathbf{H}_t$.

BiLSTMs inherently capture sequential dependencies in both forward and backward directions, naturally modeling positional relationships within sequences. This allows them to effectively replace positional embeddings in Transformer-based models, as the BiLSTM's recurrent nature already preserves order information through its hidden states. 

\textbf{Conformer Block} 
To capture both global and local inter-frame spatial dependencies, a stack of Conformer blocks is applied after the BiLSTM layer. The Conformer\cite{gulati2020conformer} is a convolution-augmented Transformer that integrates multi-head self-attention with convolutional modules to model long-range dependencies while preserving sensitivity to local patterns. Let $h^{(l)} \in \mathbb{R}^{T \times 2H}$ be the input to the $l$-th Conformer layer. The output $h^{(l+1)}$ is computed through the following residual computations:
\begin{align}
    h^{(l,1)} &= h^{(l)} + \frac{1}{2} \cdot \mathrm{FFN}(\mathrm{LayerNorm}(h^{(l)})) \\
    h^{(l,2)} &= h^{(l,1)} + \mathrm{Conv}(\mathrm{LayerNorm}(h^{(l,1)})) \\
    h^{(l,3)} &= h^{(l,2)} + \mathrm{MHSA}(\mathrm{LayerNorm}(h^{(l,2)})) \\
    h^{(l+1)} &= h^{(l,3)} + \frac{1}{2} \cdot \mathrm{FFN}(\mathrm{LayerNorm}(h^{(l,3)}))
\end{align}

Here, $\mathrm{MHSA}$ denotes the Multi-Head Self-Attention mechanism defined as:
\begin{align}
    \mathrm{MHSA}(Q, K, V) = \mathrm{Concat}(\text{head}_1, \dots, \text{head}_h) W^O
\end{align}
where each attention head is computed as:
\begin{align}
    \text{head}_i = \mathrm{Attention}(QW_i^Q, KW_i^K, VW_i^V)
\end{align}
where $Q$, $K$, and $V$ denote the query, key, and value matrices, obtained via linear projections of the input. $W_i^Q$, $W_i^K$, and $W_i^V$ are learnable projection matrices for the $i$-th head, and $W^O$ is the output projection matrix. 

In this Conformer block architecture, each module adopts residual connections and layer normalization for stable optimization. The convolution-attention fusion in the Conformer further enriches feature representations by integrating local and global information.

\textbf{Attention Pooling}
Following the Conformer blocks, attention pooling is applied to aggregate features across the sequence. Given the output \( h^{(l+1)} \in \mathbb{R}^{T \times 2H} \), the attention weights \( \mathbf{w} \) are computed as:

\begin{equation}
    \mathbf{w} = \text{Softmax}(\mathrm{Linear}(h^{(l+1)}))
\end{equation}
and the final pooled feature representation \( h_{p} \) is calculated as:

\begin{equation}
    h_p = \sum_{t=1}^{T} \mathbf{w}_t \cdot h^{(l+1)}_t
\end{equation}

This pooling mechanism enables the model to focus on the most informative parts of the sequence dynamically.

\textbf{Fully Connected Classifier} Finally, the pooled feature vector is passed through a fully connected network with linear layers, batch normalization, GELU activation, and dropout. Batch normalization stabilizes training, GELU introduces non-linearity, and dropout helps prevent overfitting, enabling the model to generalize better.

\textbf{Loss Function and Optimizer}
To guide the model training, we use the cross-entropy loss function to quantify the difference between predicted and true labels. For optimization, we employ the Adam with Weight Decay (AdamW) algorithm, which combines momentum and weight decay to enhance convergence and reduce overfitting.

\section{Experimental Evaluation}
In this section, the learning frameworks are implemented in Python 3.12.5 using PyTorch 2.4.1. Experiments are conducted on NVIDIA GeForce RTX 4070 Laptop GPU (CUDA 12.1), using approximately 870,000 samples collected from four physical channels across four traffic categories.
\subsection{Multi-Scenario Learning}
Multi-scenario learning enables models to learn and generalize across diverse environments. In our experiment, the model is trained simultaneously on physical channel data collected under 11 distinct wireless channel quality levels. To avoid data leakage, we preserve the temporal order when splitting the dataset, rather than using random shuffling as is common in traditional machine learning, since temporally adjacent samples are often highly correlated. 

\begin{table}[b]
    \centering
    \caption{Multi-Scenario Learning Performance of BiLCNet}
    \label{tab:evaluation}
    \begin{tabular}{>{\centering\arraybackslash}m{0.8cm} >{\centering\arraybackslash}m{1.8cm} cccc}
        \toprule
        Index & User Behavior & AC & PR & RC & F1 \\
        \midrule
        (1) & Call &  & 75.24\% & 87.44\% & 80.88\% \\
        (2) & Meeting &  & 88.65\% & 89.46\% & 89.05\% \\
        (3) & Upload &  & 98.81\% & 97.11\% & 97.95\% \\
        (4) & Download &  & 98.41\% & 94.04\% & 96.18\% \\
        \midrule
        \multicolumn{2}{c}{\textbf{Overall metrics}} & \textbf{93.93\%} & \textbf{90.28\%} & \textbf{92.01\%} & \textbf{91.01\%} \\
        \bottomrule
    \end{tabular}
\end{table}

We adopt accuracy (AC), precision (PR), recall (RC), and F1-score (F1) as evaluation metrics to assess the classification performance. Table~\ref{tab:evaluation} summarizes the results of BiLCNet across different traffic types. The model performs particularly well on \textit{Upload} and \textit{Download} traffic, which can be attributed to the distinct and stable feature patterns of these traffic types. In contrast, the results on \textit{Call} and \textit{Meeting} are relatively lower, possibly due to their lower occurrence or higher feature similarity. Nevertheless, the model still achieves a high overall accuracy, demonstrating its strong capability in multi-scenario traffic classification.

\begin{table}[t]
    \centering
    \footnotesize
    \setlength{\tabcolsep}{5pt} % 缩小列间距
    \caption{Performance Comparison of Different Models in Multi-Scenario Learning}
    \label{tab:model_comparison}
    \begin{tabular}{c c c c c c}
        \toprule
        Category & Models & AC & PR & RC & F1 \\
        \midrule
        \multirow{3}{*}{ML} 
            & LR & 91.32\% & 87.50\% & 89.30\% & 88.25\% \\
            & CART & 91.32\% & 87.03\% & 87.66\% & 87.32\% \\
            & RF & 92.71\% & 88.74\% & 90.12\% & 90.12\% \\
        \midrule
        \multirow{3}{*}{DL} 
            & LSTM & 93.40\% & 89.84\% & 91.40\% & 90.53\% \\
            & Transformer & 93.45\% & 90.11\% & 90.86\% & 90.45\% \\
            & BiLCNet (ours) & \textbf{93.93\%} & \textbf{90.28\%} & \textbf{92.01\%} & \textbf{91.01\%} \\
        \bottomrule
    \end{tabular}
\end{table}

\begin{figure}[b]
    \centering
    \includegraphics[width=1\linewidth]{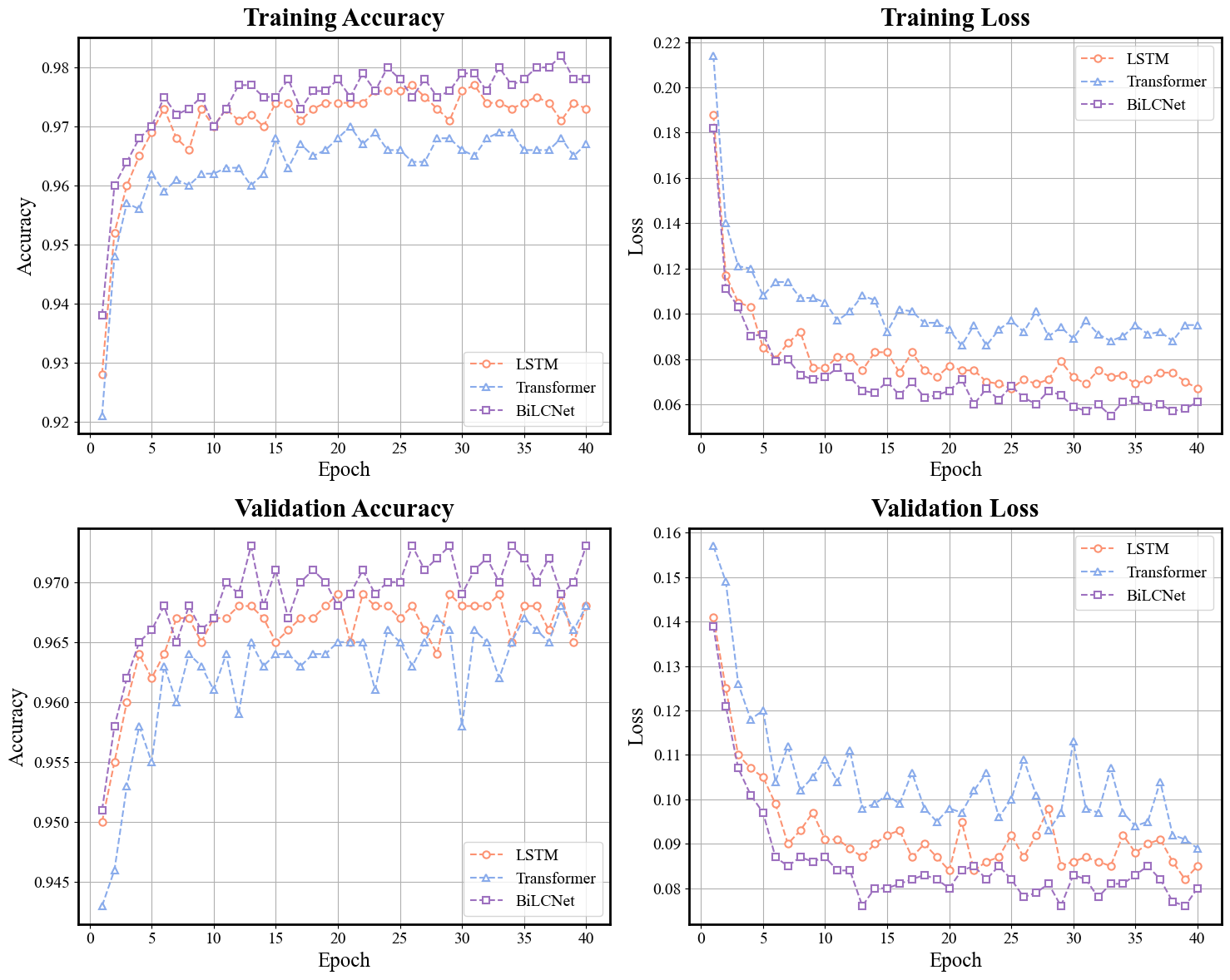}
    \caption{Deep Learning Model Comparison: Metric Evolution Across Epochs}
    \label{fig:evaluation}
\end{figure}

To further validate our model's accuracy, Table~\ref{tab:model_comparison} compares its performance with commonly used machine learning (including Logistic Regression (LR), Classification and Regression Trees (CART), and Random Forest (RF)) and deep learning models in a multi-scenario learning context. Overall, BiLCNet consistently outperforms all others across all metrics. Although the improvements appear marginal in percentage terms, they represent significant progress given the large dataset scale. Additionally, all deep learning models are trained under identical parameter settings, with 25 training epochs and early stopping. The superior performance of the hybrid model over standalone LSTM and Transformer highlights the effectiveness of combining temporal modeling with attention mechanisms to capture complex traffic patterns.

Fig.~\ref{fig:evaluation} evaluates 40-epoch training trajectories of the three deep learning architectures under multi-scenario conditions, with the vertical axis magnified for clearer comparison. The results demonstrate that the Transformer exhibits significant instability during training, while the traditional LSTM shows better stability but with an evident performance ceiling. In contrast, BiLCNet maintains superior accuracy across both training and validation phases, while also achieving the fastest convergence speed during initial training epochs.
\subsection{Zero-shot transfer learning}
\begin{figure}[h]
    \centering
    \includegraphics[width=1\linewidth]{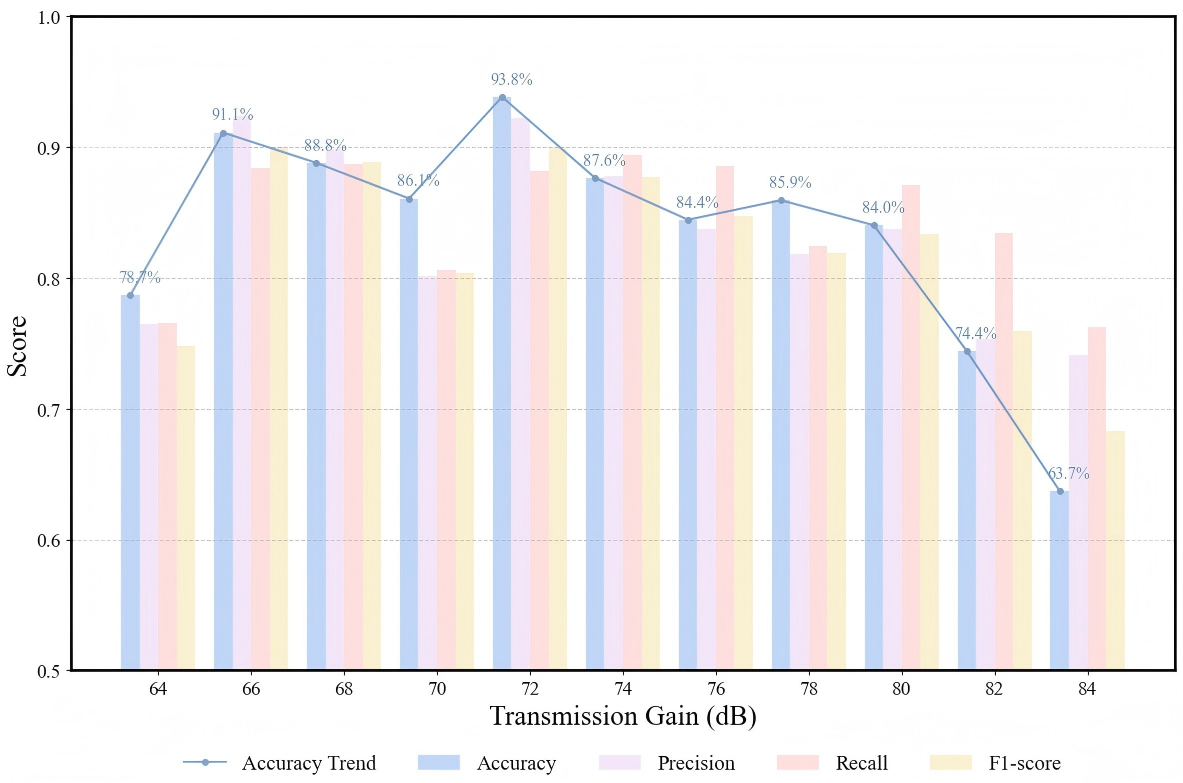}
    \caption{Zero-Shot Transfer Learning Performance of BiLCNet}
    \label{fig:zero-shot}
\end{figure}

Zero-shot transfer learning aims to evaluate a model’s generalization capability by transferring it from known tasks to unseen ones. In this study, we partition scenarios based on the transmission gain of wireless access networks, using data from each of the 11 gain settings as the test set while training the model on the remaining data. As shown in Fig.~\ref{fig:zero-shot}, each bar represents the model’s performance on a test set from an unseen gain setting, with the vertical axis scaled from 0.5 to 1 to highlight the variations.

The experimental results show that BiLCNet achieves an average accuracy of 83.4\% across the tested transmission gain spectrum, demonstrating robust overall performance. Additionally, The model exhibits a characteristic “low-high-low” performance trend, with optimal accuracy observed in the 66-80 dB range, indicating strong transfer learning capability within this interval. However, the performance degrades at both gain extremes. This suggests that the model's zero-shot transfer learning requires training samples from diverse channel conditions, where both high and low-quality channel knowledge are essential for adapting to new environments. These findings highlight opportunities to enhance model robustness across wider signal strength variations for real-world deployment.
\section{Conclusion}
In this paper, we first present a preprocessing pipeline for feature extraction from 5G SA physical channel data, collected through experimental platform simulations of real-world communication environments. We then introduce BiLCNet, a novel hybrid model for encrypted traffic classification, designed to jointly capture both temporal and spatial dependencies within channel feature matrices by combining BiLSTM, convolution and Transformer to exploit their complementary strengths. The model achieves state-of-the-art performance across multiple scenarios and demonstrates strong generalization in zero-shot learning. The proposed approach shows significant promise for real-world deployment in intelligent traffic management systems, with potential for extension to multi-device environments and diverse application scenarios.

\bibliographystyle{IEEEtran}
\bibliography{main}

\end{document}